\documentclass[sigconf]{acmart}

\copyrightyear{2026}
\acmYear{2026}
\setcopyright{cc}
\setcctype{by}
\acmConference[MM '26]{Proceedings of the 34th ACM International Conference on Multimedia}{November 10--14, 2026}{Rio de Janeiro, Brazil}
\acmBooktitle{Proceedings of the 34th ACM International Conference on Multimedia (MM '26), November 10--14, 2026, Rio de Janeiro, Brazil}
\acmDOI{10.1145/3767308.3835909}
\acmISBN{979-8-4007-2213-4/2026/11}
\AtBeginDocument{%
  }

\usepackage{multirow}

\begin{document}

\title{UnsDrive: Towards Robust End-to-End Autonomous Driving in Unstructured Scenes}

\author{Nanxin Zeng}
\authornote{Equal contribution}
\email{zengnanxin24@mails.ucas.ac.cn}
\orcid{0009-0000-0270-9193}
\affiliation{%
  \institution{University of Chinese Academy of Sciences}
  \city{Beijing}
  \country{China}
}

\author{Ruiqi Song}
\authornotemark[1]
\email{ruiqi.song@ia.ac.cn}
\orcid{0000-0003-2261-3724}
\affiliation{%
  \institution{Institute of Automation, Chinese Academy of Sciences}
  \city{Beijing}
  \country{China}
}

\author{Xiangyu Guo}
\email{xiangyu.guo@waytous.com}
\orcid{0009-0000-9131-1610}
\affiliation{%
  \institution{Waytous Inc.}
  \city{Beijing}
  \country{China}
}

\author{Baiyong Ding}
\email{knightdby@gmail.com}
\orcid{0009-0000-5228-4250}
\affiliation{%
  \institution{Waytous Inc.}
  \city{Beijing}
  \country{China}
}

\author{Yunfeng Ai}
\authornote{Corresponding author}
\email{aiyunfeng@ucas.ac.cn}
\orcid{0000-0002-2673-1368}
\affiliation{%
  \institution{University of Chinese Academy of Sciences}
  \city{Beijing}
  \country{China}
}

\renewcommand{\shortauthors}{Zeng et al.}

\begin{abstract}
End-to-end planning has shown strong promise for autonomous driving, but most existing methods are designed for structured urban roads and generalize poorly to unstructured mining environments. In such settings, weak road structure, terrain-induced occlusions, degraded visibility, and large unobserved regions make safe planning particularly challenging. To address these challenges, we propose UnsDrive, an end-to-end planner designed for unstructured mining scenes. UnsDrive builds an unknown-aware occupancy representation that explicitly models occupied, free, and unknown space using multi-frame visibility cues, and conditions a flow-matching planner on this representation to generate multimodal future trajectories. To improve safety under partial observability, we further introduce an occupancy trajectory consistency loss and an uncertainty-aware trajectory scorer that penalize trajectories entering non-traversable or unobserved regions. We also present MineLoop, a mining-oriented closed-loop simulator for evaluating autonomous driving under irregular road geometry, degraded visibility, heavy-vehicle interactions, and mining-specific operational constraints. Experiments in both open-loop and closed-loop settings show that UnsDrive consistently outperforms strong baselines in trajectory accuracy, collision avoidance, and long-horizon driving robustness. These results demonstrate the value of explicit unknown-space reasoning for autonomous driving in unstructured mining environments.
\end{abstract}

\begin{CCSXML}
<ccs2012>
   <concept>
       <concept_id>10010147.10010178.10010199.10010204</concept_id>
       <concept_desc>Computing methodologies~Robotic planning</concept_desc>
       <concept_significance>500</concept_significance>
       </concept>
   <concept>
       <concept_id>10010147.10010178.10010199.10010201</concept_id>
       <concept_desc>Computing methodologies~Planning under uncertainty</concept_desc>
       <concept_significance>300</concept_significance>
       </concept>
 </ccs2012>
\end{CCSXML}

\ccsdesc[500]{Computing methodologies~Robotic planning}
\ccsdesc[300]{Computing methodologies~Planning under uncertainty}

\keywords{Unstructured Scenes, Closed-Loop Simulation, End-to-End Planning}

\begin{teaserfigure}
  \includegraphics[width=\textwidth]{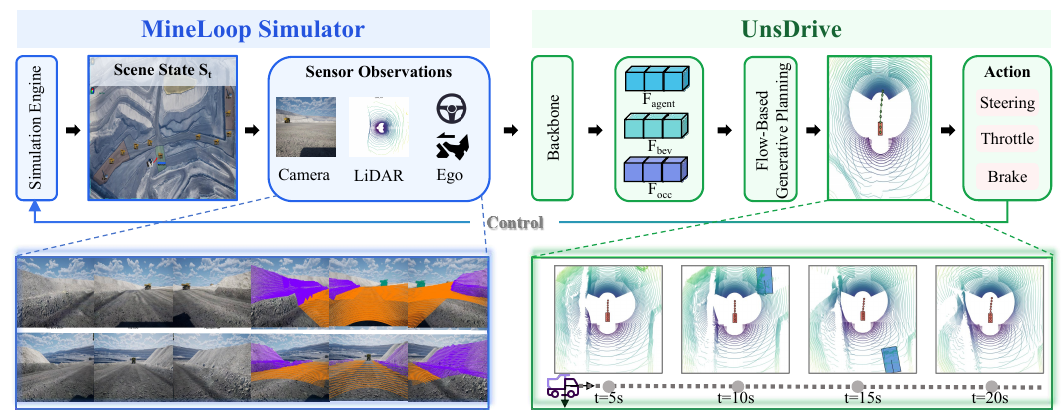}
  \caption{\textbf{Closed-loop pipeline of UnsDrive in MineLoop.} MineLoop provides synchronized camera, LiDAR, and ego-state observations through sensor simulation. UnsDrive encodes these inputs into agent, BEV, and occupancy features, which condition a flow-based planner to generate future trajectories. The planned trajectories are then converted into control actions and fed back to the simulator for closed-loop evaluation in unstructured mining environments.}
  \Description{Diagram showing the closed-loop interaction between the MineLoop simulator and UnsDrive. The simulator generates scene states and sensor observations, including camera, LiDAR, and ego information. UnsDrive encodes these inputs into agent, BEV, and occupancy features, applies a flow-based generative planner, and outputs control actions such as steering, throttle, and brake, which are fed back to the simulator.}
  \label{fig:teaser}
\end{teaserfigure}

\maketitle

\section{Introduction}
Recent autonomous driving research has increasingly shifted from modular pipelines with separately designed perception, prediction, and planning components to end-to-end frameworks that map sensor observations directly to future ego motion or driving plans~\cite{yurtsever2020survey,zhao2025survey}. By optimizing the driving system holistically, end-to-end methods reduce error accumulation across modules and enable planning-oriented representation learning. More recently, generative end-to-end planners have shown particular promise. Diffusion and flow models can produce multiple candidate trajectories that are consistent with the scene and capture the inherently multimodal nature of driving decisions~\cite{xing2025goalflow}.

However, the vast majority of these efforts have been developed and evaluated in structured urban environments, where well-maintained roads, clear lane markings, HD maps, and relatively predictable traffic patterns provide strong structural priors. When we turn to autonomous haulage in open-pit mining areas, a domain with significant industrial demand, two fundamental challenges arise that existing approaches fail to address adequately~\cite{min2024autonomous,atakishiyev2024explainable}.

First, the unique characteristics of mining environments pose severe difficulties for both perception and prediction. Unlike urban roads, mining haul roads are unstructured and lack lane markings, curbs, and other geometric cues that urban methods heavily rely on. The road topology changes frequently due to ongoing excavation and dumping activities. Heavy-duty vehicles such as mining trucks and excavators exhibit drastically different dynamics from passenger cars, and environmental factors including dust, vibration, and extreme lighting conditions further degrade sensor observations~\cite{haghighizadeh2024comprehensive}. Moreover, the observable area is often limited by terrain occlusion from pits, ramps, and stockpiles, resulting in large unknown regions that must be explicitly reasoned about rather than naively treated as free space. Existing end-to-end methods, designed with urban structural assumptions, struggle to handle these compounded challenges.

Second, there is a notable lack of closed-loop evaluation frameworks tailored to mining scenarios. Existing autonomous driving benchmarks and closed-loop simulators are predominantly built around structured urban or highway environments, encoding assumptions about road networks, traffic rules, and agent behavioral models that do not transfer to mining operations~\cite{jia2024bench2drive,dauner2024navsim}. Open-loop evaluation, while convenient, is particularly inadequate for mining autonomy: the absence of structural priors means that compounding errors, distributional shift, and the need for reactive decision-making in the presence of large unknown regions become far more pronounced than in urban settings~\cite{li2023survey}. Without a closed-loop evaluation framework that reflects mine-site geometry and mining-specific operational logic, it is difficult to reliably assess whether a proposed method truly works in practice.

To address these two challenges in a unified manner, we tackle the problem from both the method side and the evaluation side, aiming to establish a complete framework for mining-specific autonomous driving, from algorithm design to closed-loop evaluation.
On the method side, we propose UnsDrive, an end-to-end planner specifically designed for unstructured mining environments. As illustrated in Fig.~\ref{fig:teaser}, UnsDrive introduces an unknown-aware occupancy representation that explicitly distinguishes observed free space, occupied space, and unobserved regions, which is critical in mining scenes where terrain occlusion is pervasive. Built upon conditional flow matching (CFM)~\cite{lipman2022flow}, UnsDrive generates multimodal future trajectories by integrating a learned velocity field from diverse noise initializations, where the linear probability path provides a simpler training objective and more stable transport than diffusion-based planners. To further improve planning reliability under partial observability, we design an occupancy trajectory consistency (OTC) loss that penalizes trajectories inconsistent with the predicted occupancy representation, particularly those entering non-traversable or unknown regions. We further develop a trajectory scoring module that evaluates candidate trajectories against this occupancy representation, enabling safer and more reliable trajectory selection in cluttered and partially observable scenes.

On the evaluation side, we develop MineLoop, a closed-loop simulator for open-pit mining. As shown in the left part of Fig.~\ref{fig:teaser}, MineLoop digitally reconstructs mining scenarios from UAV survey data and performs closed-loop rollouts with reactive agent models calibrated to heavy-vehicle dynamics and mining operational rules. It supports configurable conditions such as dust, visibility, and road surface changes, and provides mining-specific metrics spanning safety, efficiency, and operational compliance. By enabling closed-loop evaluation under mining-specific conditions and operational logic, MineLoop helps bridge the gap between offline development and real-world deployment.

In summary, our contributions are as follows:
\begin{itemize}
    \item We propose UnsDrive, an end-to-end planner featuring an unknown-aware occupancy representation, an occupancy trajectory consistency (OTC) loss, and an uncertainty-aware trajectory scoring module, collectively addressing the unique planning challenges of unstructured mining environments.
    \item We develop MineLoop, a closed-loop simulator with mining-specific scenarios, agents, and operational logic, enabling closed-loop evaluation of autonomous driving systems in open-pit mines.
    \item Through extensive open-loop experiments and closed-loop evaluation in MineLoop, we show that UnsDrive significantly outperforms adapted urban-driving baselines in the mining domain.
\end{itemize}

\section{Related Work}
\subsection{End-to-End Autonomous Driving}
Conventional autonomous driving systems typically follow a modular perception--prediction--planning--control pipeline, which offers interpretability but is prone to interface mismatch and long-horizon error accumulation~\cite{dong2025end}. End-to-end learning alleviates these issues by jointly optimizing the mapping from raw observations to driving actions or motion plans.

Recent progress in this area increasingly favors planning-oriented end-to-end systems that predict trajectories, waypoints, occupancy, or cost maps from structured scene representations, especially bird's-eye-view (BEV) features. TransFuser~\cite{chitta2022transfuser} improves robustness through Transformer-based multi-sensor fusion, while ReasonNet~\cite{shao2023reasonnet} enhances temporal and global reasoning for more coherent long-horizon behavior. Several efforts further pursue efficiency and scalability through sparse computation and unified Transformer-based modeling~\cite{jiang2023vad,zhu2025sparsead,sun2025sparsedrive,jia2025drivetransformer}.

Beyond discriminative approaches, generative and multimodal paradigms are also introduced to end-to-end driving. DiffusionDrive~\cite{liao2025diffusiondrive} formulates motion planning as a diffusion process for multimodal trajectory generation, while OpenEMMA~\cite{xing2025openemma} and OpenDriveVLA~\cite{zhou2026opendrivevla} use multimodal models to improve waypoint generation, semantic reasoning, and instruction understanding.

\subsection{Autonomous Driving in Unstructured Environments}
Autonomous driving in unstructured environments requires reasoning over free space, traversability, and terrain geometry under weak lane, semantic, and geometric priors~\cite{min2024autonomous,wang2024survey}. Existing methods predominantly follow modular pipelines, employing search-based, sampling-based, or optimization-based planners for terrain-constrained motion generation~\cite{wang2024survey,guo2023survey,li2023trajectory,xiong2021optimized}. To compensate for the absence of lane structure, prior work has turned to dense spatial representations such as traversability maps, free-space segmentation, and occupancy grids~\cite{min2024autonomous,wang2024survey,ahn2022vision,rasib2021pixel}, highlighting the value of geometry-aware dense representations over sparse semantic abstractions in unstructured settings.

Despite these advances, most methods remain modular or rely on hand-crafted abstractions, making them vulnerable to perception noise and long-horizon error accumulation~\cite{min2024autonomous}. While end-to-end alternatives based on imitation learning and reinforcement learning have begun to emerge~\cite{guo2023survey,ahn2022vision,zhu2020off}, to the best of our knowledge, generative planning remains underexplored in unstructured driving.

\section{UnsDrive: An End-to-End Planner for Unstructured Mining Scenes}

\subsection{System Overview}

\begin{figure*}
  \includegraphics[width=\textwidth]{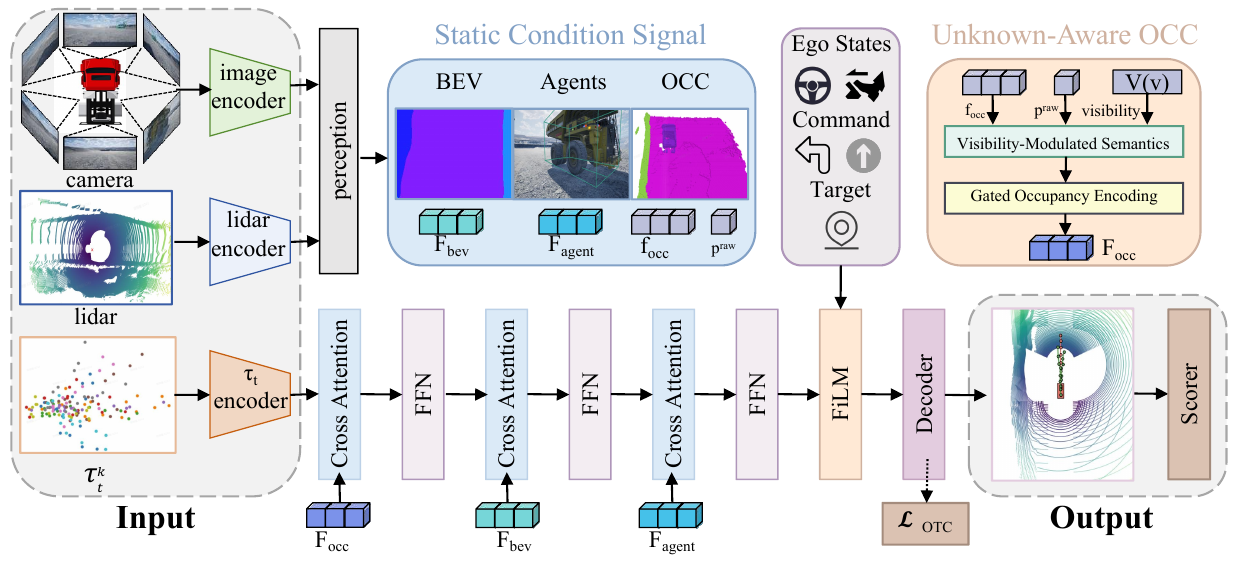}
  \caption{\textbf{Overview of UnsDrive.} Multi-view images and LiDAR are encoded into BEV, agent, and occupancy features. An unknown-aware occupancy module refines occupancy cues with visibility information. Conditioned on these features together with ego states, command, and target point, a flow-based planner generates multimodal future trajectories.} 
  \Description{Overview figure illustrating the UnsDrive end-to-end planning framework: multi-modal perception (images and LiDAR) produces BEV, agent, and unknown-aware occupancy features; a conditional flow-matching planner attends to these scene conditions and outputs candidate future ego trajectories, which are ranked by safety- and uncertainty-aware scoring.}
  \label{fig:Overview2}
\end{figure*}

UnsDrive is an end-to-end framework for autonomous driving in unstructured mining environments, consisting of a perception representation module and a generative planning module. Given multi-view camera images, LiDAR observations, the ego state, a high-level navigation command, and a target point, UnsDrive outputs a safe future ego trajectory in the BEV plane over the next \(T\) time steps. As illustrated in Fig.~\ref{fig:Overview2}, the framework first constructs a dense and uncertainty-aware scene representation from multi-modal observations, and then conditions a flow-based planner on this representation to generate multimodal trajectory candidates. We describe these two modules in detail in Sections~\ref{sec:unstructured} and~\ref{sec:planning}, respectively.

\subsection{General Representation for Unstructured Scenes}
\label{sec:unstructured}

Open-pit mining requires a scene representation that captures geometry, semantics, observability, and vertical structure. Methods developed for urban driving often collapse 3-D occupancy into BEV, but such a reduction removes height information that is critical for modeling ramps, stockpiles, and highwalls. Moreover, raw occupancy predictions do not distinguish free space from regions that remain unobserved due to occlusion, dust, or limited sensor range. To address these challenges, UnsDrive constructs an unknown-aware volumetric representation with explicit visibility modeling for unstructured mining scenes.

We use TransFuser~\cite{chitta2022transfuser} to encode multi-view images and LiDAR observations into a BEV feature map \(F_{\mathrm{bev}}\) and an agent feature \(F_{\mathrm{agent}}\). We further adopt SurroundOcc~\cite{wei2023surroundocc} to derive a volumetric representation from multi-view image features, including a latent voxel feature volume \(f_{\mathrm{occ}}(v)\) and a per-voxel semantic occupancy distribution \(p^{\mathrm{raw}}(v)\). These outputs form the basis of our unknown-aware occupancy representation. We next estimate visibility, calibrate semantics, and convert the resulting 3-D occupancy volume into occupancy tokens for downstream planning.

\noindent\textbf{Multi-frame visibility estimation.}
To quantify observability without introducing additional learnable parameters, we estimate voxel-wise visibility directly from LiDAR geometry. Specifically, LiDAR sweeps from \(N\) consecutive frames are transformed into the current ego frame and ray-traced through the voxel grid. For each frame, a voxel is labeled as occupied if any ray terminates at it, free if at least one ray passes through it and none terminates there, and unobserved otherwise, with occupied taking precedence over free. Let \(n_{\mathrm{obs}}(v)\) denote the number of frames in which voxel \(v\) is observed. To alleviate sparsity, we aggregate observations across frames and apply a one-voxel dilation only to confirmed free voxels, thereby filling small unobserved gaps while preserving occupied boundaries. The resulting visibility score is defined as:
\begin{equation}
V(v)=\frac{n_{\mathrm{obs}}(v)}{N}
\label{eq:vis}
\end{equation}

\noindent\textbf{Visibility-modulated semantics.}
Given the visibility estimate, we use it to modulate semantic occupancy and introduce an explicit unknown class. For each voxel, the original semantic distribution is scaled by \(V(v)\), while the remaining probability mass is assigned to the unknown class:
\begin{equation}
P_k(v)=V(v)\,p_k^{\mathrm{raw}}(v) \qquad
P_{\mathrm{unk}}(v)=1-V(v)
\label{eq:vis-mod}
\end{equation}
where \(k\) indexes the original semantic classes predicted by the occupancy branch. For planning, the non-traversable probability is obtained by summing over predefined obstacle classes and projecting the result to BEV via max pooling along the vertical axis. The unknown probability is projected in the same way to preserve observability information.

\noindent\textbf{Gated occupancy encoding.}
While the modulated distribution captures semantic uncertainty, it remains separate from the learned occupancy features. To combine both sources of information, we adopt a gated residual design to fuse the occupancy feature \(f_{\mathrm{occ}}(v)\) with the calibrated semantic-visibility evidence \(P(v)\). Specifically, the occupancy feature is first projected into the token space to obtain \(\hat{f}_{\mathrm{occ}}(v)\), while the unknown probability \(P_{\mathrm{unk}}(v)\) is mapped by a lightweight MLP to an uncertainty embedding \(u(v)\). A channel-wise gating vector is then computed from these two components to adaptively balance occupancy information and uncertainty:
\begin{equation}
g(v)=\sigma\!\bigl(W_g[\hat{f}_{\mathrm{occ}}(v);u(v)] + b_g\bigr)
\label{eq:gate}
\end{equation}
where \(W_g\) and \(b_g\) are the learnable parameters of the gating layer. In addition, the full semantic probability vector \(P(v)\) is linearly projected to the same token dimension. The final occupancy token is defined as:
\begin{equation}
F_{\mathrm{occ}}(v)=g(v)\odot \hat{f}_{\mathrm{occ}}(v) + \bigl(1-g(v)\bigr)\odot u(v) + W_P P(v)
\label{eq:occ-token}
\end{equation}
where \(W_P\) denotes the linear projection applied to \(P(v)\). This design preserves geometric context while explicitly encoding semantics and uncertainty.

\noindent\textbf{Volumetric tokenization.}
The resulting per-voxel tokens still form a dense 3-D grid. To preserve height information while reducing token count, we avoid collapsing the voxel grid into a 2-D BEV map. Instead, we apply a lightweight 1-D convolution along the vertical axis to compress the \(Z\) dimension. The resulting volume is then flattened into a token sequence with a learnable 3-D positional encoding. Together with the BEV features \(F_{\mathrm{bev}}\) and the agent-state features \(F_{\mathrm{agent}}\), these vertically compressed 3-D occupancy tokens form the final scene condition \(\mathcal{C}\) for the generative planner.

\subsection{Flow-Based Generative Planning}
\label{sec:planning}

Given the unknown-aware scene condition \(\mathcal{C}\) derived in Section~\ref{sec:unstructured}, the planner must reason about both multimodal future evolution and safety under partial observability. In unstructured mining environments, the lack of lane topology, traffic rules, and other structured priors leads to a highly multimodal distribution over feasible future motions. Under such conditions, deterministic regression-based planners often average across distinct modes and therefore produce unrealistic or invalid trajectories. To address this issue, we formulate planning as conditional generative modeling and adopt CFM as the core planner. This formulation naturally supports multimodal trajectory generation while allowing the planning process to be conditioned on uncertainty-aware scene representations.

\noindent\textbf{Trajectory representation and flow matching.}
We represent the future plan as a BEV trajectory \(\tau\) with \(T\) waypoints in the ego-centric frame. To learn the conditional distribution of feasible futures, we employ CFM. During training, a noise trajectory \(z\) is sampled and paired with the ground-truth trajectory \(\tau\). An intermediate state \(\tau_t\) is then constructed by linear interpolation between \(z\) and \(\tau\) at time \(t\in[0,1]\), and the target velocity \(u_t\) is defined as the transport direction from \(z\) to \(\tau\). Conditioned on \(\mathcal{C}\), the planner predicts the velocity field \(v_\theta(\tau_t,t,\mathcal{C})\) and is optimized with:
\begin{equation}
\mathcal{L}_{\mathrm{cfm}}=\mathbb{E}\bigl[\|v_\theta(\tau_t,t,\mathcal{C})-u_t\|^2\bigr]
\end{equation}

\begin{figure*}[t]
  \centering
  \includegraphics[width=\textwidth]{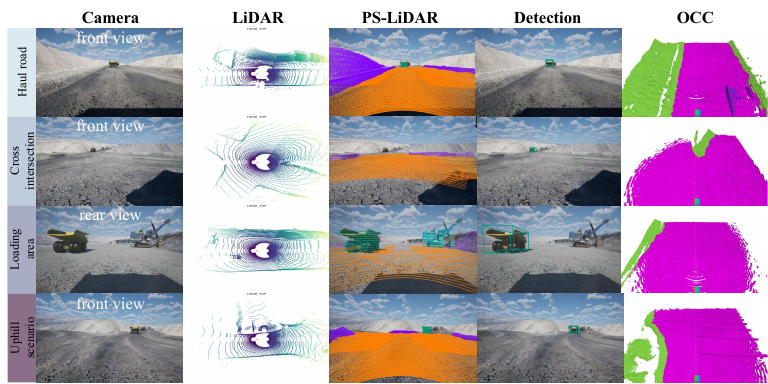}
  \caption{\textbf{MineLoop scenarios.} Representative scenes under different mining conditions.}
  \Description{Representative scenes under different mining conditions.}
  \label{fig:mineloop_vis}
\end{figure*}

\noindent\textbf{Flow-state encoding and scene fusion.}
To estimate the conditional velocity field, the intermediate flow state must be fused with the heterogeneous scene condition. We first embed \(\tau_t\) into \(D\)-dimensional waypoint tokens using an MLP, and encode the flow time \(t\) with sinusoidal positional encoding added to the waypoint tokens. We then fuse the scene condition \(\mathcal{C}=(F_{\mathrm{occ}}, F_{\mathrm{bev}}, F_{\mathrm{agent}})\) through a cascaded cross-attention design rather than direct concatenation. Specifically, the planner first attends to the unknown-aware 3-D occupancy tokens \(F_{\mathrm{occ}}\) to inject safety and uncertainty cues, then to the BEV feature map \(F_{\mathrm{bev}}\) to incorporate global spatial context, and finally to the agent feature \(F_{\mathrm{agent}}\) to capture dynamic interactions. This safety-first ordering prioritizes occupancy-aware reasoning while preserving scene-level and interaction information.

\noindent\textbf{Navigation condition injection.}
After scene fusion, navigation intent is injected through feature-wise linear modulation (FiLM). The navigation command, goal point, and ego state are encoded as \(e_{\mathrm{nav}}\), \(e_{\mathrm{goal}}\), and \(e_{\mathrm{ego}}=\mathrm{MLP}([v; a; \omega])\), respectively, and combined into a unified command embedding \(e_{\mathrm{cmd}} = e_{\mathrm{nav}} + e_{\mathrm{goal}} + e_{\mathrm{ego}}\). For the \(l\)-th fusion block, the scene-fused trajectory feature can be modulated as follows:
\begin{equation}
h_l = \gamma\!\left(e_{\mathrm{cmd}}\right)\odot \tilde{h}_l + \beta\!\left(e_{\mathrm{cmd}}\right)
\end{equation}
where \(\tilde{h}_l\) and \(h_l\) denote the input and output features of the \(l\)-th FiLM block, respectively, and \(\odot\) denotes element-wise multiplication. We stack \(L\) such fusion blocks for iterative refinement, and decode the final feature with an MLP to predict the velocity field \(v_{\theta}(\tau_t,t,\mathcal{C})\).

\noindent\textbf{Occupancy trajectory consistency regularization.}
Although CFM captures the multimodal distribution of expert trajectories, it does not explicitly prevent generated plans from entering non-traversable or unobserved regions. To couple generation with safety reasoning, we introduce an occupancy trajectory consistency loss \(\mathcal{L}_{\mathrm{OTC}}\). For a predicted trajectory, the risk of waypoint \((x_i,y_i)\) is computed by bilinearly sampling the BEV non-traversable probability \(P_{\mathrm{occ}}\) and unknown probability \(P_{\mathrm{unk}}\), both obtained by max pooling the 3-D occupancy prediction along the vertical axis. The waypoint risk is defined as:
\begin{equation}
r_i = P_{\mathrm{occ}}(x_i, y_i) + \alpha P_{\mathrm{unk}}(x_i, y_i)
\end{equation}
where \(\alpha \in (0,1)\) controls the penalty imposed on unknown regions. The consistency loss is:
\begin{equation}
\mathcal{L}_{\mathrm{OTC}} = \frac{1}{T}\sum_{i=1}^{T}\max(r_i - \varepsilon, 0)
\end{equation}
where \(\varepsilon\) is a safety tolerance.

\noindent\textbf{Candidate generation and uncertainty-aware scoring.}
The flow-based planner generates multiple feasible futures by integrating the learned velocity field from different Gaussian initializations. We sample \(K\) candidate trajectories and assign each of them an uncertainty-aware score composed of a learned trajectory-quality term, a non-traversable occupancy penalty derived from \(P_{\mathrm{occ}}\), and an unknown-region penalty derived from \(P_{\mathrm{unk}}\). This scoring mechanism enables the planner to balance motion quality against traversability and observability when selecting among multimodal futures. At test time, the candidate with the highest score is selected as the final plan. To train the scoring head, we use a margin-based ranking objective:
\begin{equation}
\mathcal{L}_{\mathrm{score}} = \sum_{k \neq k_{\mathrm{ref}}}\max\left(0,\, s_k - s_{k_{\mathrm{ref}}} + m\right)
\end{equation}
where \(s_k\) denotes the score of the \(k\)-th candidate, \(k_{\mathrm{ref}}\) denotes the candidate closest to the expert trajectory, and \(m\) is the ranking margin.

\noindent\textbf{Overall training objective.}
Finally, we train the full model end-to-end with the loss as follows:
\begin{equation}
\mathcal{L} = \mathcal{L}_{\mathrm{cfm}} + \lambda_{\mathrm{occ}} \mathcal{L}_{\mathrm{occ}} + \lambda_{\mathrm{OTC}} \mathcal{L}_{\mathrm{OTC}} + \lambda_{\mathrm{score}} \mathcal{L}_{\mathrm{score}}
\end{equation}
where \(\mathcal{L}_{\mathrm{occ}}\) denotes the occupancy semantic segmentation loss, and \(\lambda_{\mathrm{occ}}\), \(\lambda_{\mathrm{OTC}}\), and \(\lambda_{\mathrm{score}}\) are balancing weights. This joint objective unifies occupancy estimation, uncertainty-aware safety regularization, and multimodal generative planning within a single end-to-end training framework.

\section{MineLoop: A Closed-Loop Simulator for Unstructured Mining Scenes}
\label{sec:closedloop}

\subsection{Overview}
Open-pit mining differs from urban driving in weak road semantics, heavy-vehicle dynamics, and degraded visibility under dust, fog, rain, and night illumination. These properties require both a standardized closed-loop simulator interface and mining-specific scene construction.

We therefore build MineLoop, a closed-loop simulator for unstructured mining scenes. MineLoop adopts a simulator--policy interface with synchronized sensing, control, logging, and rollout management, and combines it with mining-oriented virtual scenes, heavy-duty agents, and operational workflows such as loading, hauling, and dumping. This design enables closed-loop benchmarking and data collection in mining environments.

\subsection{Closed-Loop Architecture}
As shown in Fig.~\ref{fig:teaser}, MineLoop implements a closed-loop simulator--policy pipeline similar to CARLA~\cite{dosovitskiy2017carla}. At each step, the simulator updates the scene state and renders synchronized multimodal observations, including camera images, LiDAR point clouds, and ego states. These observations are encoded by the backbone into agent, BEV, and occupancy features, which are then fed to UnsDrive.

UnsDrive predicts future motion and generates control actions with a flow-based planner. The resulting steering, throttle, and brake commands are fed back to MineLoop, where the ego vehicle is propagated by the dynamics model and surrounding agents are updated by the behavior engine. This closed-loop interaction enables long-horizon evaluation under mining conditions. Rollouts terminate on collision, prolonged off-road behavior, route completion, or timeout. 

\subsection{Scene Construction and Supervision}
Fig.~\ref{fig:mineloop_vis} shows representative MineLoop scenes under different mining conditions. Following the scenario-engineering pipeline of PMWorld~\cite{ai2023pmworld}, MineLoop is built on mining-oriented virtual scenes reconstructed from on-site surveying and UAV mapping, capturing terrain geometry together with mining-specific semantic regions such as haul roads, ramps, intersections, loading zones, dumping areas, and temporary work zones. These regions define traversability, route sampling, and off-road detection.

MineLoop models mining-specific agents including trucks, excavators, loaders, vehicles, and static obstacles, with heavy-duty dimensions and task-aware motion constraints. A hybrid behavior engine supports common mining interactions such as car-following, overtaking, yielding, queuing, and loading-hauling-dumping cycles. Traffic density, visibility, weather, and obstacle placement can be varied to generate both routine and stress-test cases.

To match the onboard stack, MineLoop simulates time-stamped camera images, LiDAR point clouds, and control signals. Six surround-view cameras run at 2~Hz with resolution \(1920\times1080\), and four Ouster OS2-128 LiDARs run at 10~Hz. The ego vehicle is propagated by the same 16-DoF dynamics model as PMWorld~\cite{ai2023pmworld}, while surrounding agents are updated by the behavior engine. The sensor stack also supports perturbations such as visibility degradation, range dropout, and illumination variation.

For occupancy-based learning and evaluation, MineLoop generates engine-consistent supervision directly from simulator geometry and agent states: occupied, free, and unknown regions follow from scene geometry and visibility, and future occupancy labels are obtained by rolling the simulator forward. MineLoop also records trajectories, control traces, and event-level metrics, including collision, off-road, route completion, progress, and minimum time-to-collision.

\section{Experiments}
\begin{table*}[t]
\setlength{\tabcolsep}{11.2pt}
\centering
\caption{\textbf{Comparison with representative methods under open-loop evaluation on the mining dataset.}}
\label{tab:mining_openloop}
\begin{tabular}{l|c|cccc|cccc}
\toprule
\multirow{2}{*}{Method} & \multirow{2}{*}{Aux. Sup.} &
\multicolumn{4}{c|}{L2 (m) $\downarrow$} & \multicolumn{4}{c}{Collision Rate (\%) $\downarrow$}\\
& & 1s & 2s & 3s & Avg & 1s & 2s & 3s & Avg \\
\midrule
OccWorld~\cite{zheng2024occworld} & Occ
& 0.63 & 1.43 & 3.16 & 1.74
& 0.29 & 0.63 & 2.95 & 1.29 \\
OccNet~\cite{tong2023scene} & Occ \& Map \& Agent
& 1.46 & 2.59 & 3.73 & 2.59
& 0.37 & 0.87 & 3.28 & 1.51 \\
SparseWorld~\cite{dang2026sparseworld} & Occ
& 0.43 & 0.77 & 1.35 & 0.85
& 0.07 & 0.18 & 0.75 & 0.33 \\
DiffusionDrive~\cite{liao2025diffusiondrive} & BEV \& Map \& Agent
& 0.47 & 0.84 & 1.42 & 0.91
& 0.07 & 0.19 & 0.83 & 0.36 \\
\midrule
\textbf{UnsDrive (ours)} & Occ \& BEV \& Agent & \textbf{0.29} & \textbf{0.41} & \textbf{0.61} & \textbf{0.44}
& \textbf{0.04} & \textbf{0.06} & \textbf{0.19} & \textbf{0.10} \\
\bottomrule
\end{tabular}%
\end{table*}

\begin{table}[t]
\setlength{\tabcolsep}{8pt}
\caption{\textbf{Closed-loop evaluation on MineLoop.}}
\centering
\begin{tabular}{l|cc}
\toprule
\multirow{2}{*}{Method} & \multicolumn{2}{c}{Closed-loop Metric} \\
& Driving Score $\uparrow$ & Success Rate (\%) $\uparrow$ \\
\midrule
OccWorld~\cite{zheng2024occworld} & 71.48 & 47.26 \\
OccNet~\cite{tong2023scene} & 63.23 & 43.69\\
SparseWorld~\cite{dang2026sparseworld} & 77.36 & 58.27\\
DiffusionDrive~\cite{liao2025diffusiondrive} & 75.14 & 53.14\\
 \midrule
 \textbf{UnsDrive (ours)} & \textbf{84.25} & \textbf{68.56} \\
\bottomrule
\end{tabular}
\label{tab:mining_closedloop}
\end{table}

\subsection{Experimental Setup}
\noindent\textbf{Dataset.}
We evaluate our method in both open-loop and closed-loop settings, using a simulation dataset collected in MineLoop and a closed-loop benchmark built on top of it.

For open-loop evaluation, the dataset is annotated under a nuScenes-style planning protocol and provides synchronized multi-view images, LiDAR point clouds, ego states, navigation commands, future ego trajectories, and occupancy labels, supervising both scene understanding and trajectory planning. It comprises 89 scenes with 34,229 frames, including 11,054 keyframes, and covers haul roads, ramps, T-junctions, crossroads, loading areas, dumping zones, and temporary work regions across the full loading-hauling-dumping workflow. These scenarios exhibit complex terrain geometry, irregular lane structures, dynamic agents, and task-oriented navigation demands, making them well suited for evaluating planning in unstructured mining environments.

For closed-loop evaluation, we build a mining benchmark on MineLoop following a Bench2Drive-inspired protocol~\cite{jia2024bench2drive}, adopting route-based interactive evaluation under feedback execution so that the planner influences subsequent observations and agent interactions throughout the rollout. It focuses on mining-specific tasks such as haul-road following, obstacle avoidance, loading-area departure, dumping-area approach, and temporary work-zone traversal. Compared with urban-driving benchmarks, these tasks place stronger emphasis on long-horizon stability, command compliance, and safe motion generation under weak structural priors, providing a challenging testbed for autonomous driving in open-pit mines.

\noindent\textbf{Implementation Details.}
For open-loop evaluation, the flow planner is trained from scratch for 10 epochs, while the scene prediction module is initialized from pre-trained weights. 
We use AdamW with a batch size of 48. 
The learning rates are \(3\times10^{-4}\) for the flow and perception modules, and \(3\times10^{-5}\) for the scene prediction module. 
For closed-loop evaluation on MineLoop, the model is trained for 100 epochs with learning rates of \(6\times10^{-4}\) for the flow module and \(6\times10^{-5}\) for the scene prediction module.

\noindent\textbf{Evaluation Metrics.}
For open-loop evaluation, we use trajectory L2 error and collision rate, both reported at 1 s, 2 s, 3 s, and averaged over the horizon. 
L2 error measures the Euclidean distance between predicted and ground-truth ego trajectories, while collision rate evaluates whether the predicted trajectory intersects with dynamic or static obstacles. 
Lower values are better.

For closed-loop evaluation on MineLoop, we use Driving Score and Success Rate. 
Driving Score reflects overall driving quality by accounting for route completion and penalty-triggering events, while Success Rate measures the percentage of episodes that successfully reach the destination. 
Higher values are better.

\subsection{Comparison Experiments}
\noindent\textbf{Open-loop Results.}
Table~\ref{tab:mining_openloop} presents the open-loop comparison results on the mining dataset. Overall, all baseline methods exhibit noticeable performance degradation in mining scenarios, indicating a substantial domain gap between conventional urban-driving benchmarks and mining environments. This degradation is reflected in both larger trajectory prediction errors and higher collision rates, which can be attributed to irregular road topology, weak lane structure, heavy-duty vehicles, and complex agent interactions in mining scenes.

Despite these challenges, UnsDrive achieves the best performance across all reported metrics. Specifically, UnsDrive obtains L2 errors of 0.29\,m, 0.41\,m, and 0.61\,m at 1\,s, 2\,s, and 3\,s, respectively, with the lowest average L2 error of 0.44\,m. In terms of safety, UnsDrive also achieves the lowest collision rates of 0.04\%, 0.06\%, and 0.19\%, yielding the best average collision rate of 0.10\%. Compared with the strongest baseline, SparseWorld, UnsDrive reduces the average L2 error from 0.85\,m to 0.44\,m and the average collision rate from 0.33\% to 0.10\%, demonstrating more accurate and safer trajectory planning in complex mining conditions.

\begin{figure*}
  \includegraphics[width=\textwidth]{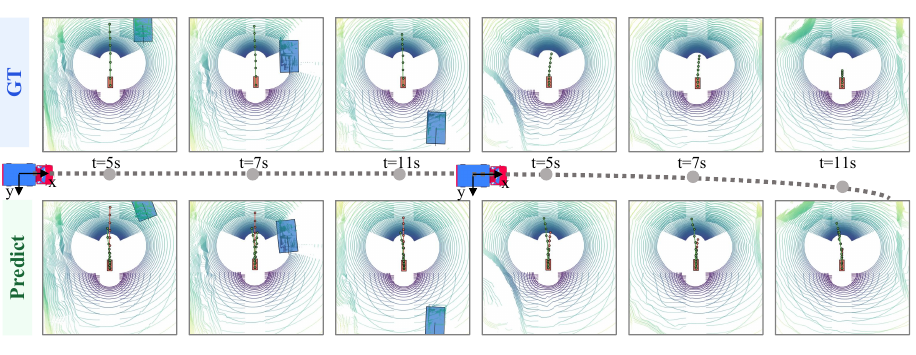}
  \caption{\textbf{Qualitative results in simulation.} Temporal trajectory visualizations in two representative mining scenarios. For each scenario, the top row shows the ground-truth future trajectory, and the bottom row shows the predicted multimodal trajectories at different time steps. The red trajectory denotes the best prediction, while the two green trajectories denote the worst predictions.}
  \Description{Qualitative simulation visualization with two scenarios, Passing and Turn Right, each shown at three timesteps. Ground-truth occupancy evolution is displayed in the top row and predictions in the bottom row. Predicted frames also show three trajectory candidates, consisting of the highest-scored trajectory and two lowest-scored alternatives.}
  \label{fig:qualitative_simulation}
\end{figure*}

\noindent\textbf{Closed-loop Results.}
Table~\ref{tab:mining_closedloop} reports the closed-loop evaluation results on the MineLoop benchmark. UnsDrive achieves the best performance on both metrics, reaching a driving score of 84.25 and a success rate of 68.56\%, ahead of the strongest baseline SparseWorld by 6.89 and 10.29 points, respectively.

These results show that the advantage of UnsDrive remains consistent under closed-loop execution, where even small planning errors can accumulate and affect future observations and interactions. The higher driving score suggests better overall driving quality and trajectory stability throughout the rollout, while the improved success rate indicates stronger robustness in completing long-horizon driving tasks. This advantage may be attributed to the unknown-aware occupancy representation and the flow-based planner conditioned on it, which together enable explicit reasoning about non-traversable and unobserved space during trajectory generation. Such capability is particularly important in open-pit mines, where road layouts are weakly structured and driving conditions vary across operational regions.

\subsection{Ablation Study}
We conduct ablation studies on three key components of the proposed framework, namely unknown-aware occupancy conditioning (UA-OCC), the OTC loss, and the uncertainty-aware scorer, under both open-loop and closed-loop settings.

\noindent\textbf{Open-loop ablation of key components.}
As shown in Table~\ref{tab:ablation_unknown_otc_score}, UA-OCC reduces the average L2 error by about 26\% and the collision rate by roughly one third over the baseline, suggesting that explicitly modeling unknown regions provides more reliable spatial context for trajectory generation under weak structural priors. Adding OTC cuts the average L2 error by another 18\% and the collision rate by 46\%, indicating that enforcing occupancy trajectory consistency suppresses physically inconsistent trajectories. With the uncertainty-aware scorer, the full model reduces the average L2 error by more than 50\% and the collision rate by nearly 75\% relative to the baseline, showing that uncertainty-aware candidate selection is complementary to the training-time regularization from UA-OCC and OTC.

\noindent\textbf{Closed-loop ablation of key components.}
Table~\ref{tab:closedloop_ablation_components} presents the closed-loop ablation results. Consistent with the open-loop observations, adding UA-OCC leads to clear improvements in closed-loop driving performance, with both Driving Score and Success Rate increasing noticeably over the baseline. When all components are enabled, the proposed method achieves the best closed-loop performance by a clear margin, demonstrating that the proposed design improves not only offline trajectory quality but also long-horizon execution under feedback.

Overall, the ablation results indicate that the three components play complementary roles: UA-OCC improves scene-aware conditioning, OTC enhances occupancy trajectory consistency during training, and the uncertainty-aware scorer improves candidate selection during inference.

\begin{table}[t]
\setlength{\tabcolsep}{3.2pt}
\centering
\caption{\textbf{Ablation on key components.} }
\label{tab:ablation_unknown_otc_score}
\begin{tabular}{c c c|cccc|cccc}
\toprule
\multicolumn{3}{c|}{Components} &
\multicolumn{4}{c|}{L2 (m) \(\downarrow\)} &
\multicolumn{4}{c}{Collision Rate (\%) \(\downarrow\)} \\
UA & OTC & Scorer &
1s & 2s & 3s & Avg &
1s & 2s & 3s & Avg \\
\midrule
\(\times\) & \(\times\) & \(\times\) &
0.43 & 0.94 & 1.52 & 0.96 &
0.12 & 0.31 & 0.73 & 0.39 \\
\(\checkmark\) & \(\times\) & \(\times\) &
0.35 & 0.62 & 1.17 & 0.71 &
0.07 & 0.24 & 0.46 & 0.26 \\
\(\checkmark\) & \(\checkmark\) & \(\times\) &
0.32 & 0.48 & 0.93 & 0.58 &
0.07 & 0.13 & 0.23 & 0.14 \\
\(\checkmark\) & \(\checkmark\) & \(\checkmark\) &
0.29 & 0.41 & 0.61 & 0.44 &
0.04 & 0.06 & 0.19 & 0.10 \\
\bottomrule
\end{tabular}
\end{table}

\begin{table}[t]
\centering
\setlength{\tabcolsep}{5.5pt}
\caption{\textbf{Ablation on closed-loop components.}}
\label{tab:closedloop_ablation_components}
\begin{tabular}{c c c|cc}
\toprule
\multicolumn{3}{c|}{Components} & \multicolumn{2}{c}{Closed-loop Metric \(\uparrow\)} \\
UA-OCC & OTC & Scorer & Driving Score & Success Rate (\%) \\
\midrule
\(\times\) & \(\times\) & \(\times\) & 76.14 & 55.12 \\
\(\checkmark\) & \(\times\) & \(\times\) & 79.37 & 63.89 \\
\(\checkmark\) & \(\checkmark\) & \(\times\) & 82.54 & 65.61 \\
\(\checkmark\) & \(\checkmark\) & \(\checkmark\) & 84.25 & 68.56 \\
\bottomrule
\end{tabular}
\end{table}

\subsection{Qualitative Results in Simulation}

Fig.~\ref{fig:qualitative_simulation} presents qualitative simulation results for two representative scenarios, passing and turn right, over multiple future time steps. For each scenario, we visualize the ground-truth future trajectory together with the predicted multimodal trajectories, including the highest-scored trajectory and two low-scored alternatives.

As shown in the figure, the highest-scored trajectory remains well aligned with the ground-truth future motion across the prediction horizon, indicating that the model can capture plausible future behavior in challenging mining scenarios. Specifically, in the turn right scenario, the highest-scored trajectory decelerates and enters the right branch with a small steering angle, consistent with the low-speed turning behavior of heavy-duty mining vehicles. In contrast, the low-scored trajectories exhibit clear deviations from the ground truth and correspond to less plausible motion patterns.

This comparison highlights that the scorer does not merely rank trajectory candidates by local preference, but effectively identifies the mode that is most consistent with the scene context. These qualitative results further demonstrate the importance of uncertainty-aware scoring for selecting reliable trajectories from multimodal predictions.

\section{Conclusion}
This work addresses end-to-end autonomous driving in unstructured mining environments, where weak scene structure, degraded visibility, and large unknown regions make reliable planning particularly challenging. We contribute an occupancy-aware generative planning paradigm together with a mining-oriented closed-loop simulator, and experiments in both open-loop and closed-loop settings demonstrate consistent gains in prediction quality, safety, and driving robustness over strong baselines. These results highlight the importance of explicitly reasoning about unknown space and perception uncertainty for autonomous driving in unstructured scenes. A key limitation is that all experiments are conducted in simulation; validating UnsDrive on physical mining trucks remains important future work. Beyond deployment, it would also be valuable to study tighter coupling between uncertainty-aware planning and downstream control.

\begin{acks}
This work was supported by the Key Research and Development Program of Shaanxi Province (2024CY2-GJHX-49) and the Key Research and Development Program of Xinjiang Uyghur Autonomous Region (Grant No. 2025B01001).
\end{acks}

\bibliographystyle{ACM-Reference-Format}
\balance
\bibliography{occdrive}

@article{ahn2022vision,
  title={Vision-based autonomous driving for unstructured environments using imitation learning},
  author={Ahn, Joonwoo and Kim, Minsoo and Park, Jaeheung},
  journal={arXiv preprint arXiv:2202.10002},
  year={2022}
}

@article{ai2023pmworld,
  title={PMWorld: A parallel testing platform for autonomous driving in mines},
  author={Ai, Yunfeng and Liu, Yuhang and Gao, Yu and Zhao, Chen and Cheng, Xiang and Han, Jinpeng and Tian, Bin and Chen, Long and Wang, Fei-Yue},
  journal={IEEE Transactions on Intelligent Vehicles},
  volume={9},
  number={1},
  pages={1402--1411},
  year={2023},
  publisher={IEEE}
}

@article{atakishiyev2024explainable,
  title={Explainable artificial intelligence for autonomous driving: A comprehensive overview and field guide for future research directions},
  author={Atakishiyev, Shahin and Salameh, Mohammad and Yao, Hengshuai and Goebel, Randy},
  journal={IEEE Access},
  volume={12},
  pages={101603--101625},
  year={2024},
  publisher={IEEE}
}

@article{chitta2022transfuser,
  title={Transfuser: Imitation with transformer-based sensor fusion for autonomous driving},
  author={Chitta, Kashyap and Prakash, Aditya and Jaeger, Bernhard and Yu, Zehao and Renz, Katrin and Geiger, Andreas},
  journal={IEEE transactions on pattern analysis and machine intelligence},
  volume={45},
  number={11},
  pages={12878--12895},
  year={2022},
  publisher={IEEE}
}

@inproceedings{dang2026sparseworld,
  title={Sparseworld: A flexible, adaptive, and efficient 4d occupancy world model powered by sparse and dynamic queries},
  author={Dang, Chenxu and Liu, Haiyan and Bao, Jason and An, Pei and Tang, Xinyue and Pan, An and Ma, Jie and Sun, Bingchuan and Wang, Yan},
  booktitle={Proceedings of the AAAI Conference on Artificial Intelligence},
  volume={40},
  number={5},
  pages={3497--3505},
  year={2026}
}

@article{dauner2024navsim,
  title={Navsim: Data-driven non-reactive autonomous vehicle simulation and benchmarking},
  author={Dauner, Daniel and Hallgarten, Marcel and Li, Tianyu and Weng, Xinshuo and Huang, Zhiyu and Yang, Zetong and Li, Hongyang and Gilitschenski, Igor and Ivanovic, Boris and Pavone, Marco and others},
  journal={Advances in Neural Information Processing Systems},
  volume={37},
  pages={28706--28719},
  year={2024}
}

@article{dong2025end,
  title={End-to-End Autonomous Driving: From Classic Paradigm to Large Model Empowerment—A Comprehensive Survey},
  author={Dong, Wei and Lu, Sikai and Chen, Xinhe and Zhang, Shunyao and Liu, Qingchao and Liu, Ze and Chen, Long and Wang, Hai and Cai, Yingfeng},
  journal={IEEE Internet of Things Journal},
  volume={13},
  number={3},
  pages={3870--3898},
  year={2025},
  publisher={IEEE}
}

@inproceedings{dosovitskiy2017carla,
  title={CARLA: An open urban driving simulator},
  author={Dosovitskiy, Alexey and Ros, German and Codevilla, Felipe and Lopez, Antonio and Koltun, Vladlen},
  booktitle={Conference on robot learning},
  pages={1--16},
  year={2017},
  organization={PMLR}
}

@article{guo2023survey,
  title={A survey of trajectory planning methods for autonomous driving—Part I: Unstructured scenarios},
  author={Guo, Yuqing and Guo, Zelin and Wang, Yazhou and Yao, Danya and Li, Bai and Li, Li},
  journal={IEEE Transactions on Intelligent Vehicles},
  volume={9},
  number={9},
  pages={5407--5434},
  year={2023},
  publisher={IEEE}
}

@article{haghighizadeh2024comprehensive,
  title={Comprehensive analysis of heavy metal soil contamination in mining Environments: Impacts, monitoring Techniques, and remediation strategies},
  author={Haghighizadeh, Atoosa and Rajabi, Omid and Nezarat, Arman and Hajyani, Zahra and Haghmohammadi, Mina and Hedayatikhah, Soheila and Asl, Soheila Delnabi and Beni, Ali Aghababai},
  journal={Arabian Journal of Chemistry},
  volume={17},
  number={6},
  pages={105777},
  year={2024},
  publisher={Elsevier}
}

@article{jia2024bench2drive,
  title={Bench2drive: Towards multi-ability benchmarking of closed-loop end-to-end autonomous driving},
  author={Jia, Xiaosong and Yang, Zhenjie and Li, Qifeng and Zhang, Zhiyuan and Yan, Junchi},
  journal={Advances in Neural Information Processing Systems},
  volume={37},
  pages={819--844},
  year={2024}
}

@article{jia2025drivetransformer,
  title={Drivetransformer: Unified transformer for scalable end-to-end autonomous driving},
  author={Jia, Xiaosong and You, Junqi and Zhang, Zhiyuan and Yan, Junchi},
  journal={arXiv preprint arXiv:2503.07656},
  year={2025}
}

@inproceedings{jiang2023vad,
  title={Vad: Vectorized scene representation for efficient autonomous driving},
  author={Jiang, Bo and Chen, Shaoyu and Xu, Qing and Liao, Bencheng and Chen, Jiajie and Zhou, Helong and Zhang, Qian and Liu, Wenyu and Huang, Chang and Wang, Xinggang},
  booktitle={Proceedings of the IEEE/CVF International Conference on Computer Vision},
  pages={8340--8350},
  year={2023}
}

@article{li2023trajectory,
  title={Trajectory planning for autonomous driving in unstructured scenarios based on deep learning and quadratic optimization},
  author={Li, Han and Chen, Peng and Yu, Guizhen and Zhou, Bin and Li, Yiming and Liao, Yaping},
  journal={IEEE Transactions on Vehicular Technology},
  volume={73},
  number={4},
  pages={4886--4903},
  year={2023},
  publisher={IEEE}
}

@article{li2023survey,
  title={A survey on self-evolving autonomous driving: a perspective on data closed-loop technology},
  author={Li, Xincheng and Wang, Zhaoyi and Huang, Yanjun and Chen, Hong},
  journal={IEEE Transactions on Intelligent Vehicles},
  volume={8},
  number={11},
  pages={4613--4631},
  year={2023},
  publisher={IEEE}
}

@inproceedings{liao2025diffusiondrive,
  title={Diffusiondrive: Truncated diffusion model for end-to-end autonomous driving},
  author={Liao, Bencheng and Chen, Shaoyu and Yin, Haoran and Jiang, Bo and Wang, Cheng and Yan, Sixu and Zhang, Xinbang and Li, Xiangyu and Zhang, Ying and Zhang, Qian and others},
  booktitle={Proceedings of the Computer Vision and Pattern Recognition Conference},
  pages={12037--12047},
  year={2025}
}

@article{min2024autonomous,
  title={Autonomous driving in unstructured environments: How far have we come?},
  author={Min, Chen and Si, Shubin and Wang, Xu and Xue, Hanzhang and Jiang, Weizhong and Chen, Zitong and Li, Mengmeng and Mei, Jilin and Shang, Erke and Xiao, Zhipeng and others},
  journal={arXiv preprint arXiv:2410.07701},
  year={2024}
}

@article{rasib2021pixel,
  title={Pixel level segmentation based drivable road region detection and steering angle estimation method for autonomous driving on unstructured roads},
  author={Rasib, Marya and Butt, Muhammad Atif and Riaz, Faisal and Sulaiman, Adel and Akram, Muhammad},
  journal={IEEE Access},
  volume={9},
  pages={167855--167867},
  year={2021},
  publisher={IEEE}
}

@inproceedings{shao2023reasonnet,
  title={Reasonnet: End-to-end driving with temporal and global reasoning},
  author={Shao, Hao and Wang, Letian and Chen, Ruobing and Waslander, Steven L and Li, Hongsheng and Liu, Yu},
  booktitle={Proceedings of the IEEE/CVF conference on computer vision and pattern recognition},
  pages={13723--13733},
  year={2023}
}

@inproceedings{sun2025sparsedrive,
  title={Sparsedrive: End-to-end autonomous driving via sparse scene representation},
  author={Sun, Wenchao and Lin, Xuewu and Shi, Yining and Zhang, Chuang and Wu, Haoran and Zheng, Sifa},
  booktitle={2025 IEEE International Conference on Robotics and Automation (ICRA)},
  pages={8795--8801},
  year={2025},
  organization={IEEE}
}

@inproceedings{tong2023scene,
  title={Scene as occupancy},
  author={Tong, Wenwen and Sima, Chonghao and Wang, Tai and Chen, Li and Wu, Silei and Deng, Hanming and Gu, Yi and Lu, Lewei and Luo, Ping and Lin, Dahua and others},
  booktitle={Proceedings of the IEEE/CVF International Conference on Computer Vision},
  pages={8406--8415},
  year={2023}
}

@article{wang2024survey,
  title={A survey on path planning for autonomous ground vehicles in unstructured environments},
  author={Wang, Nan and Li, Xiang and Zhang, Kanghua and Wang, Jixin and Xie, Dongxuan},
  journal={Machines},
  volume={12},
  number={1},
  pages={31},
  year={2024},
  publisher={MDPI}
}

@inproceedings{wei2023surroundocc,
  title={Surroundocc: Multi-camera 3d occupancy prediction for autonomous driving},
  author={Wei, Yi and Zhao, Linqing and Zheng, Wenzhao and Zhu, Zheng and Zhou, Jie and Lu, Jiwen},
  booktitle={Proceedings of the IEEE/CVF International Conference on Computer Vision},
  pages={21729--21740},
  year={2023}
}

@inproceedings{xing2025openemma,
  title={Openemma: Open-source multimodal model for end-to-end autonomous driving},
  author={Xing, Shuo and Qian, Chengyuan and Wang, Yuping and Hua, Hongyuan and Tian, Kexin and Zhou, Yang and Tu, Zhengzhong},
  booktitle={Proceedings of the Winter Conference on Applications of Computer Vision},
  pages={1001--1009},
  year={2025}
}

@article{xiong2021optimized,
  title={An optimized trajectory planner and motion controller framework for autonomous driving in unstructured environments},
  author={Xiong, Lu and Fu, Zhiqiang and Zeng, Dequan and Leng, Bo},
  journal={Sensors},
  volume={21},
  number={13},
  pages={4409},
  year={2021},
  publisher={MDPI}
}

@article{yurtsever2020survey,
  title={A survey of autonomous driving: Common practices and emerging technologies},
  author={Yurtsever, Ekim and Lambert, Jacob and Carballo, Alexander and Takeda, Kazuya},
  journal={IEEE access},
  volume={8},
  pages={58443--58469},
  year={2020},
  publisher={IEEE}
}

@inproceedings{xing2025goalflow,
  title={Goalflow: Goal-driven flow matching for multimodal trajectories generation in end-to-end autonomous driving},
  author={Xing, Zebin and Zhang, Xingyu and Hu, Yang and Jiang, Bo and He, Tong and Zhang, Qian and Long, Xiaoxiao and Yin, Wei},
  booktitle={2025 IEEE/CVF Conference on Computer Vision and Pattern Recognition (CVPR)},
  pages={1602--1611},
  year={2025},
  organization={IEEE}
}

@article{zhao2025survey,
  title={A survey of autonomous driving from a deep learning perspective},
  author={Zhao, Jingyuan and Wu, Yuyan and Deng, Rui and Xu, Susu and Gao, Jinpeng and Burke, Andrew},
  journal={ACM Computing Surveys},
  volume={57},
  number={10},
  pages={1--60},
  year={2025},
  publisher={ACM New York, NY}
}

@inproceedings{zheng2024occworld,
  title={Occworld: Learning a 3d occupancy world model for autonomous driving},
  author={Zheng, Wenzhao and Chen, Weiliang and Huang, Yuanhui and Zhang, Borui and Duan, Yueqi and Lu, Jiwen},
  booktitle={European conference on computer vision},
  pages={55--72},
  year={2024},
  organization={Springer}
}

@inproceedings{zhou2026opendrivevla,
  title={Opendrivevla: Towards end-to-end autonomous driving with large vision language action model},
  author={Zhou, Xingcheng and Han, Xuyuan and Yang, Feng and Ma, Yunpu and Tresp, Volker and Knoll, Alois},
  booktitle={Proceedings of the AAAI Conference on Artificial Intelligence},
  volume={40},
  number={16},
  pages={13782--13790},
  year={2026}
}

@article{zhu2025sparsead,
  title={Sparsead: Sparse query-centric paradigm for efficient end-to-end autonomous driving},
  author={Zhu, Runwen and Zhao, Jianbo and Zhang, Diankun and Wang, Guoan and Chen, Xiwu and Zhang, Siyu and Gong, Jiahao and Zhou, Qibin and Zhang, Wenyuan and Wang, Ningzi and others},
  journal={IEEE Transactions on Artificial Intelligence},
  year={2025},
  publisher={IEEE}
}

@inproceedings{zhu2020off,
  title={Off-road autonomous vehicles traversability analysis and trajectory planning based on deep inverse reinforcement learning},
  author={Zhu, Zeyu and Li, Nan and Sun, Ruoyu and Xu, Donghao and Zhao, Huijing},
  booktitle={2020 IEEE intelligent vehicles symposium (IV)},
  pages={971--977},
  year={2020},
  organization={IEEE}
}

@article{lipman2022flow,
  title={Flow matching for generative modeling},
  author={Lipman, Yaron and Chen, Ricky TQ and Ben-Hamu, Heli and Nickel, Maximilian and Le, Matt},
  journal={arXiv preprint arXiv:2210.02747},
  year={2022}
}

@String{Computing = "Computing" }

@String{Computer = "{IEEE} Computer" }

@String{Springer = "Springer-Verlag" }

\appendix

\end{document}